\documentclass[times,art10,twocolumn,latex8]{article}
\usepackage{latex8}
\usepackage{times}
\usepackage{amsmath}
\usepackage{amsfonts}
\usepackage{amssymb}
\usepackage{graphicx}
\usepackage{epsfig}

\begin{document}

\title{Multiresolution Elastic Medical Image Registration in Standard Intensity Scale}

\author{Ula\c{s} Ba\u{g}c\i \\
The University of Nottingham\\ Collaborative Medical Image Analysis Group\\ 
Jubilee Campus, NG8 1BB, UK \\ uxb@cs.nott.ac.uk\\
\and
Li Bai\\
The University of Nottingham\\ Collaborative Medical Image Analysis Group\\ 
Jubilee Campus, NG8 1BB, UK \\ bai@cs.nott.ac.uk\\
}

\maketitle
\thispagestyle{empty}

\begin{abstract}
Medical image registration is a difficult problem. Not only a registration algorithm needs to capture both large and small scale image deformations, it also has to deal with global and local image intensity variations. In this paper we describe a new multiresolution elastic image registration method that challenges these difficulties in image registration. To capture large and small scale image deformations, we use both global and local affine transformation algorithms. To address global and local image intensity variations, we apply an image intensity standardization algorithm to correct image intensity variations. This transforms image intensities into a standard intensity scale, which allows highly accurate registration of medical images.   
\end{abstract}

\section{Introduction}
Image registration is important in many imaging applications. For example, in diagnostic imaging, there is often a need for comparing two images for disease diagnosis or longitudinal studies. There is also a frequent need for registering two images of different imaging modalities, e.g., MR and PET images.

Image registration involves the development of a reasonable transformation between a pair of images, \textit{the source} and \textit{target}, such that the similarity between the transformed source image (registered source) and target image is optimum. The similarity measure should capture both large and small scale deformations (also known as displacements), together with global and local variations of image intensities. Based on the nature of the transformation, registration methods can be categorized as \textit{rigid}, \textit{affine} and \textit{elastic}. In rigid registration, the transformation includes global rotation and/or translation parameters. For Affine registration in particular, scaling parameters are also included. Registration is considered elastic(deformable) if the transformation is able to express both global and local deformations. For surveys of image registration including nonlinear medical image registration, see~\cite{survey, medsurvey, nonmedsurvey, surveybook, nonmedsurvey2, survey2, janbook}. 

Although rigid and affine transformations are able to align images, they can only handle global deformations. In rigid registration, the recovered transformation itself has no clinical significance, however, in nonrigid registration the recovered transformation may have clinical significance~\cite{surveybook}. Since motion and deformation characteristics are necessary for quantification of changes between images, transformation should be found as accurate as possible. Except for a few studies~\cite{periaswamy, towards, rueckert, fast}, most of the elastic deformations based nonrigid registrations rely on the assumption that image intensities remain constant between images~\cite{elastic1, elastic2, nonmedsurvey, nonmedsurvey2}, which is not always true and affects the accuracy of motion and deformations obtained from the transformation.

To address this problem, a locally affine but globally smooth transformation model has been developed in the presence of intensity variations in~\cite{periaswamy}. In addition to 6 and 12 affine parameters for 2D and 3D registrations respectively, two more affine parameters are used to capture intensity variations during registration. In order to remove inefficiency and inaccuracy arising from certain circumstances, such as low-resolution images, Bayesian based importance sampling technique with the same spatially varying parameters are used in~\cite{towards}. In~\cite{rueckert}, voxel based similarity measures, such as normalized mutual information, are combined with B-spline based nonrigid transformation called free-form deformation (FFD). Since the intensity and contrast between the pre- and post-contrast enchanced images vary, voxel based similarity measures are used because it is insensitive to intensity changes as a result of contrast enchancement. However, there is a trade-off between accuracy and computation time of FFD-based method. The local flexibility and computational complexity of the local motion model is related to the resolution of B-spline control points. More control points may improve the registration accuracy, but the computation time will also increase dramatically~\cite{breastsurvey}. In~\cite{fast}, a fast elastic multidimensional intensity-based image registration with a parametric model of deformation is presented. Although adding landmarks controls the smoothness of deformation field and using a multiresolution approach for both the image and the deformation model makes registration algorithm robust and fast, global solution of the optimization function cannot be guaranteed due to manual identification of landmarks.

In this paper, we present a multiresolution elastic image registration framework on images in the standard intensity scale. The standard intensity scale is obtained by a standardisation procedure which corrects image intensity variations~\cite{udupa_std_jmri}. In the standard scale, similar intensities will mean similar tissue properties. 

The paper is organised as follows: a detailed description of elastic registration used in this study is given in Section~\ref{sec:affine}. A brief explanation of intensity standardization method is presented in Section~\ref{sec:std}. Multiresolution framework is explained in Section~\ref{sec:multi}. Experiments and promising results are given in Section~\ref{sec:results} and concluding remarks in Section~\ref{sec:conc}. 
\section{Local Affine Transformation}
\label{sec:affine}
For 2D image registration, an affine transformation has six parameters, which can be determined if the coordinates of at least three non-colinear corresponding points in the images are known~\cite{regbook}. As $\vec{v}=[x \quad y \quad 1]^T$ represents the homogeneous spatial coordinates, let $F(\vec{v},t)$ and $F(\vec{v}^*,t-1)$ be the source and target images respectively. General affine transformation between source and target image can be modelled locally as:

\begin{equation}
\label{eq:affine}
\left[ a_7 \quad a_8\right] .\left[\begin{array}{c} F(\vec{v},t)\\ 1 \end{array}\right]=F(\vec{v}^*,t-1)=F(A.\vec{v},t-1)
\end{equation}
where $t$ is the time, $a_7$ and $a_8$ are parameters describing intensity variations between source and target images and $A$ is affine transform matrix defined as:

\begin{equation}
A=
\left[ \begin{array}{ccc}
a_{1} & a_{2} & a_{3} \\
a_{4} & a_{5} & a_{6} \\
0        &  0        & 1
\end{array} \right]
\end{equation} 
In general notation of affine transformation, $a_7$ and $a_8$ controls the intensity variations between image pairs. In proposed registration algorithm since possible variations in intensities are captured by standardization method, Eq~(\ref{eq:affine}) is minimized by setting $a_7=1$ and $a_8=0$ respectively. A simple way to estimate affine matrix parameters is to minimize quadratic error function $E(A)$ which can be defined as:

\begin{equation}
\label{eq:error}
E(A) \approx  \sum_{\vec{v}\in D} \left[ F(\vec{v},t)-F(A.\vec{v},t-1)\right] ^2
\end{equation} 
where $D$ denotes a small spatial neighborhood. From a Taylor expansion of Eq~(\ref{eq:error}), we obtain linear quadratic error function to be minimized:
\begin{eqnarray}
\label{eq:linearerror}
E(A) \approx  \sum_{\vec{v}\in D} \left( f_t - (\vec{f}^T. [A-I]). \vec{v}\right) ^2 \nonumber \\
\approx  \sum_{\vec{v}\in D} \left( f_t - \vec{f}^T.A.\vec{v}+\vec{f}^T.\vec{v}\right) ^2
\end{eqnarray}
where I is 3x3 identity matrix, $\vec{f}=[f_x  \quad f_y \quad f_t]^T$ and
\begin{equation}
f_x = f_x(\vec{v},t)  \qquad f_y = f_y(\vec{v},t)  \qquad  f_t = f_t(\vec{v},t) 
\end{equation} 
are partial derivatives of image $F$ on $D$. Open expression for the gradient based constraint equation ~(\ref{eq:linearerror}) can be expressed further as:
\begin{eqnarray}
E(\vec{a})= \sum_{x,y\in D} (x.f_x.a_1+y.f_x.a_2+f_x.a_3+x.f_y.a_4\nonumber \\
+y.f_y.a_5+f_y.a_6 - f^T.\vec{v} ) ^2 \nonumber \\
\end{eqnarray} 
where $\vec{a}=[a_1\quad a_2 \quad  a_3 \quad  a_4 \quad  a_5 \quad  a_6]^T$. More compact form can be obtained by defining $\vec{\Omega}=[x.f_x \quad  y.f_x  \quad   f_x  \quad  x.f_y  \quad  y.f_y \quad,  f_y]^T$ as in~\cite{periaswamy}:
\begin{equation}
\label{eq:finalerror}
E(\vec{a})= \sum_{x,y\in D} [\vec{\Omega}^T.\vec{a}-f^T.\vec{v}]^2
\end{equation}
Quadratic error function in Eq~(\ref{eq:finalerror}) can be minimized analytically by differentiating it with respect to the unknown parameters $\vec{a}$
\begin{equation}
\label{eq:derivative}
 dE(\vec{a})/d\vec{a}=\sum_{x,y\in D}2\vec{\Omega}\left[ \vec{\Omega}^T.\vec{a}-f^T.\vec{v}\right] 
\end{equation}
Setting Eq~(\ref{eq:derivative}) equal to zero and solving for $\vec{a}$ parameters yields:
\begin{equation}
\label{eq:estimation}
\vec{a}=\left[ \sum_{x,y\in D}\vec{\Omega}\vec{\Omega}^T\right] ^{-1}\left[ \sum_{x,y\in D}\vec{\Omega}f^T.\vec{v}\right] 
\end{equation}
Since the velocity field at each image point has two components while the changes in image brightness at a point in the image plane due to motion yields only one constraint, the optical flow cannot be computed at a point in the image independently of neighboring points without introducing additional constraints~\cite{horn}. This additional constraint is based on the smoothness of parameters over domain $D$ such that neighboring points on the domain $D$ have similar velocities and the velocity field of the brightness patterns in the image varies smoothly almost everywhere. One way to express this additional constraint is to minimize the magnitude of the gradient of the optical flow velocity, which is:
\begin{equation}
\label{eq:smooth}
E_{smooth}(\vec{a})=\sum_{i=1}^{6}\lambda_i \left[ \left( \frac{\partial a_i}{\partial x}\right)^2+\left( \frac{\partial a_i}{\partial y}\right)^2\right]  
\end{equation}
where magnitude of $\lambda_i$ reflects the influence of smoothness term. Hence, the problem  is to minimize the sum of the errors in the equation~(\ref{eq:finalerror}) and~(\ref{eq:smooth}). To obtain local affine parameters $\vec{a}_i$, we can differentiate $E_{total}(\vec{a})=E(\vec{a})+E_{smooth}(\vec{a})$ and set the result to be zero~\cite{horn}. Solution for the $\vec{a}$ is~\cite{towards}:
\begin{equation}
\label{eq:solution}
\vec{a}^{n+1}=\left(\vec{\Omega}.\vec{\Omega}^T+\Lambda\right)^{-1}\left( \vec{\Omega}.\vec{f}^T.\vec{v}+\Lambda.\vec{a}_p^{n}\right)  
\end{equation}  
where $\Lambda$ is 6x6 diagonal matrix with elements $\lambda_i$, and $\vec{a}_p^n$ is the component-wise average of $\vec{a}$ over domain $D$. Starting with the initial guess $(\vec{a}^0)$  \footnote{$\vec{a}^0$ can be obtained through the same equation by putting $\Lambda=null$,~\cite{periaswamy}} , at each step the next local affine paramters are computed and resultant system of linear equations are solved accordingly.

\section{Standardization of MR Image Intensity Scale}
\label{sec:std}
Since MR image intensities do not have a fixed tissue meaning in image scale even within the same protocol, for the same body region, for images obtained on the same scanner, for the same patient, there is a strong need to transform image scale into standard intensity scale in order so that for the same MR protocol and body region, similar intensities will have similar tissue meaning~\cite{udupa_std_jmri}.  

Standardization is based on mapping image intensity histograms into a standard histogram. The method consists of two steps called training and transformation. In the training step, a set of images of the same body region are given as input to "learn" histogram-specific parameters, called landmarks. In the transformation step, any given image is standardized with estimated histogram-specific landmarks obtained from the training step.

\subsection{Methods}
Based on the study~\cite{udupa_std}, image histograms of the same body region are always of the same type. There are different types of histograms, unimodal, bimodal and generalization of both. Since most of the protocols studied in~\cite{udupa_std} produce bimodal histograms, bimodal histogram distribution is used to extract histogram specific landmarks. In bimodal histograms, one of the histogram specific landmarks is mode($\mu$) representing main foreground object in the image, as depicted in Figure 1. 

\begin{figure}[h]
\label{im:bimodal}
\begin{center}
\caption{Location of the histogram specific landmarks, $m_1$=minimum gray value, $m_2$=maximum gray value}
\includegraphics[scale=0.6]{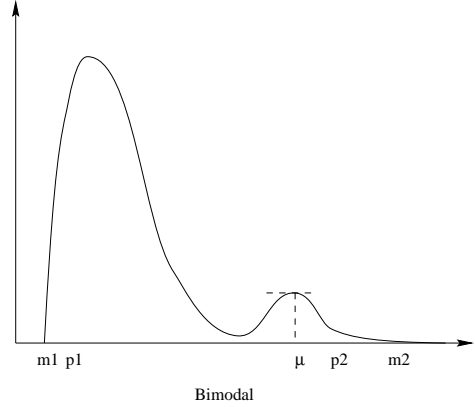} 
\end{center}
\end{figure}

Other histogram specific landmarks denoted by $p_1$ and $p_2$ are extracted according to range of intensity of interest (IOI). Landmarks $p_1$ and $p_2$ are defined according to  minimum and maximum percentile values, $pc_1$ and $pc_2$, that are used to select IOI. 

In the training step, for image $j$, the landmarks $p_{1j}$, $p_{2j}$ and $\mu_j$ obtained from the histogram of each image are mapped to the standard scale by mapping intensities from $[p_{1j}, p_{2j}]$ to $[s_1, s_2]$ where $s_1$ and $s_2$ are minimum and maximum intensities on the standard scale respectively. The formula for mapping $x\in [p_{1j}, p_{2j}]$ to $x'$ is the following~\cite{udupa_std}.
\begin{equation}
\label{eq:mapping}
x'=s_1 + \frac{x-p_{1j}}{p_{2j}-p_{1j}}(s_2-s_1)
\end{equation}

Figure 2 shows two separate linear mappings, the first from $[p_{1i},\mu_i]$ to $[s_1,\mu_s]$ and the second from $[\mu_i, p_{2i}]$ to $[\mu_s, s_2]$. Overall mapping, $\tau_i(x)$, from $[m_{1i}, m_{2i}]$ to $[s_{1i}', s_{2i}']$  can be summarized as follows:

\begin{equation}
\label{eq:finalmapping}
\tau_i(x)= \left\{ \begin{array}{ll}
\lceil \mu_s+ (x-\mu_i)\left( \frac{s_1-\mu_s}{p_{1i}-\mu_i}\right) \rceil & \textrm{if $m_{1i} \leq  x \leq \mu_i$}\\
\lceil \mu_s+ (x-\mu_i)\left( \frac{s_2-\mu_s}{p_{2i}-\mu_i}\right) \rceil & \textrm{if $\mu_i \leq x \leq m_{2i}$}
\end{array} \right.
\end{equation} 
where $\lceil . \rceil$ converts any number y$\in \Re$ into closest integer Y such that Y $\geq y$ or $\leq y$. Further details can be found in~\cite{udupa_std}.

\begin{figure}[h]
\begin{center}
\caption{The intensity mapping function for the transformation step}
\includegraphics[scale=0.7]{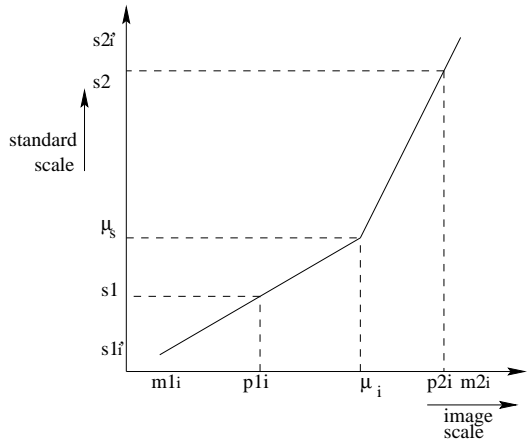} 
\end{center}
\end{figure}

\begin{figure}[h]
\begin{center}
\caption{Histograms before and after standardization}
 \includegraphics[scale=0.50]{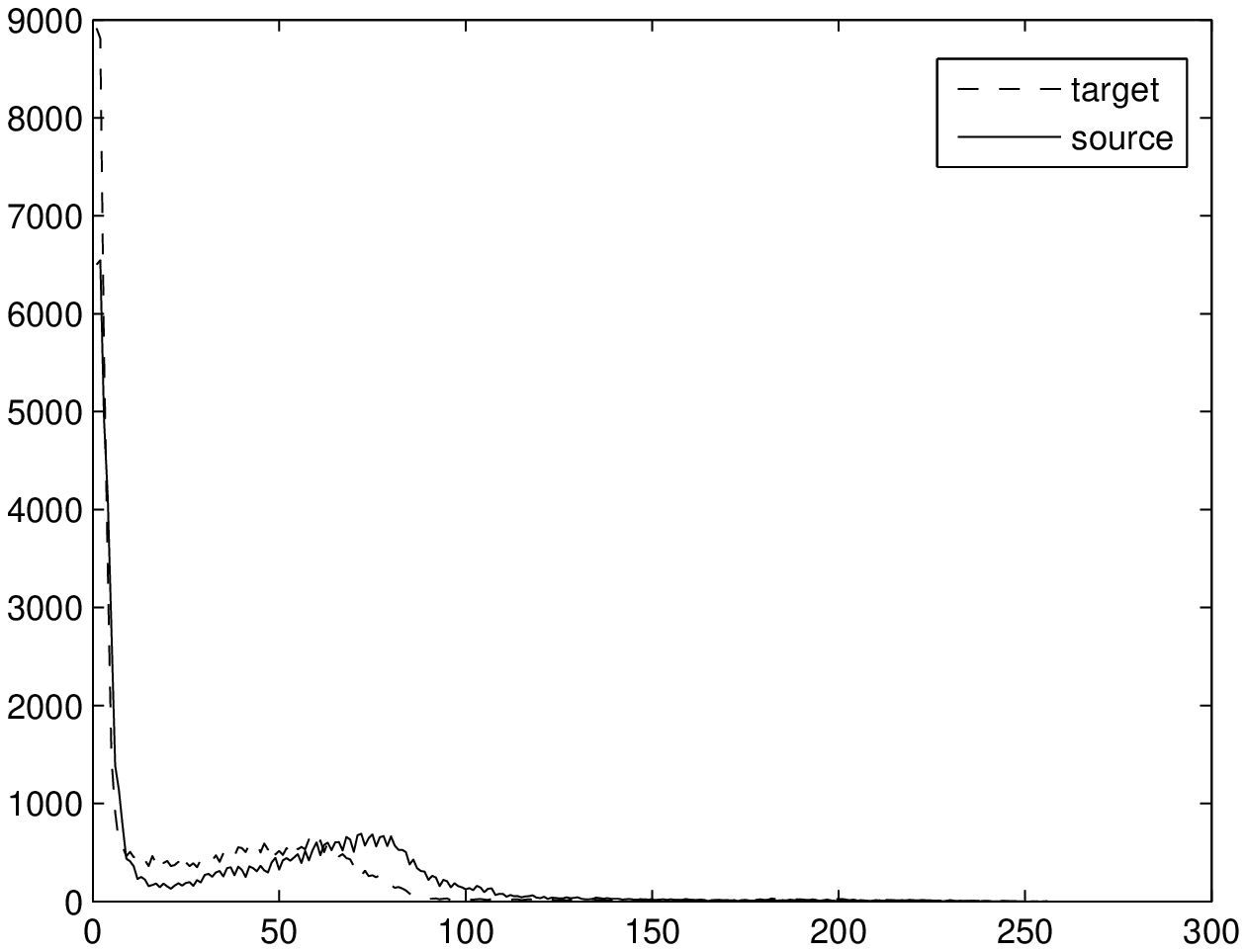}
 \includegraphics[scale=0.50]{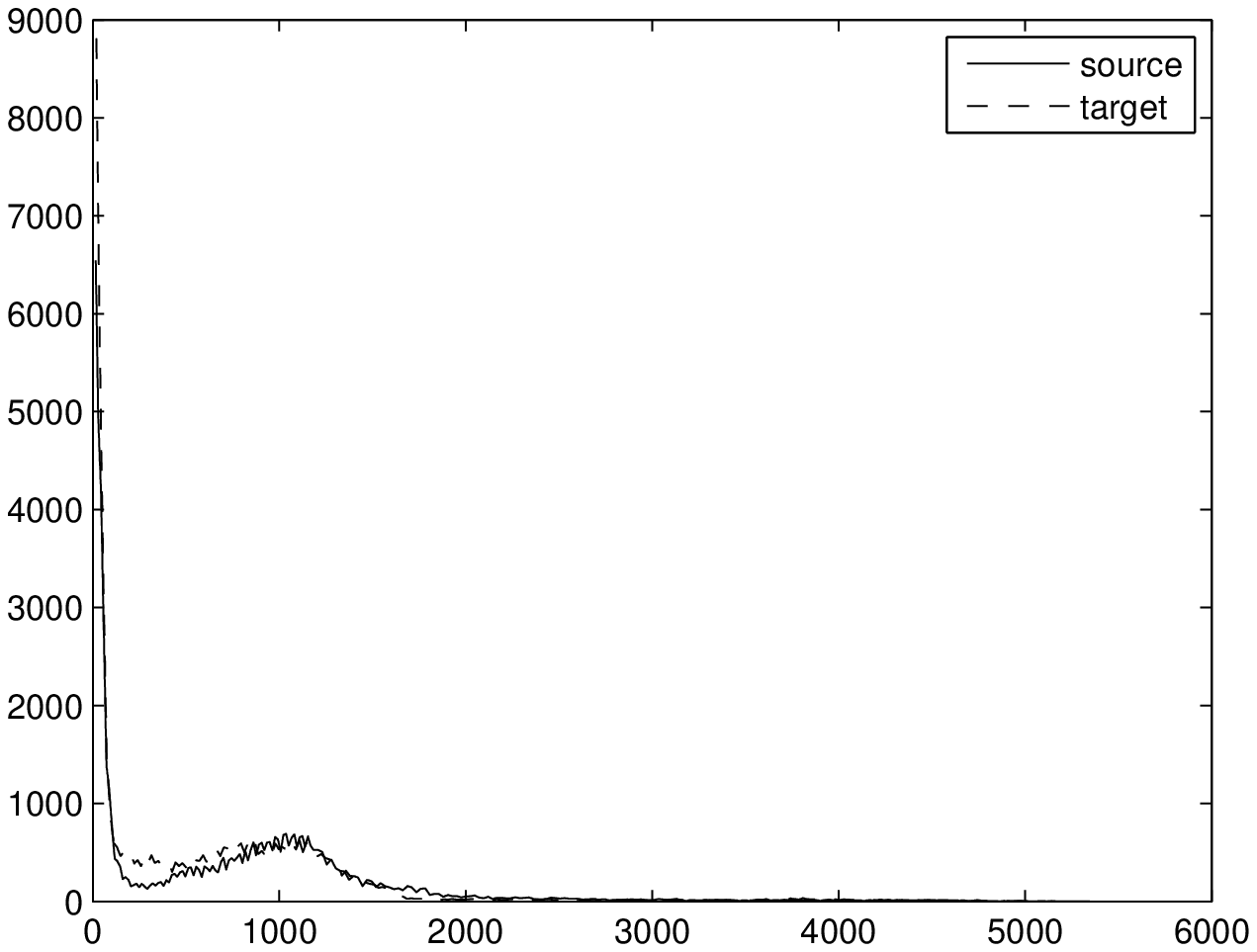}  
\end{center}
\end{figure}

\subsection{Choosing the Standardization Parameters}
Based on the experiments in~\cite{udupa_std,udupa_std_jmri}, minimum and maximum percentile values are set to $pc_{1}=0$ and $pc_2=99.8$ respectively. On the standard scale, $s_1$ and $s_2$ are set to $s_1=1$ and $s_2=4095$. Figure 3 shows the histograms before intensity mapping and after intensity mapping respectively. It is easily seen that histograms are more similar in shape, location and distribution after standardization than before. It means that intensities have tissue-specific meaning after the standardization.    

Images in the first row of Figure 4 shows a source and target images in image scale. In the second row on the same figure, the same slices are displayed after standardization using the parameters defined above.

\begin{figure}[h]
\begin{center}
\caption{images in non-standard(1st row) and standard scale(2nd row)}
\includegraphics[scale=0.40]{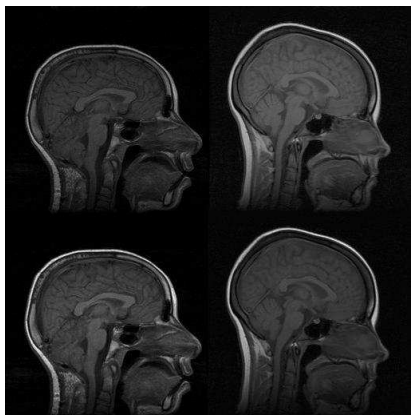} 
\end{center}
\end{figure} 


\section{Multiresolution}
\label{sec:multi}
Image registration can be highly nonlinear and therefore many iterations may be required to reach a solution. An important method to reduce the amount of computational cost and deal with nonlinearities is to use a multilevel(multiresolution) image pyramid. Multilevel continuation is well established for optimization problems and systems of non-linear equations~\cite{jan}.  As common to many other nonrigid registration algorithms, the registration method we use includes two steps. After global registration has been done in the first step, locally affine globally smooth elastic registration on standard scale is performed. In order to achieve low computational cost and accelerate the registration process, coarse-to-fine strategy is used. Global affine registration is first performed at the coarsest level where convergence is fast because there are few data. The initial condition at the coarsest scale is arbitrary. Moreover, it is likely that the criterion to optimize has a reduced number of local optima; this is due to a loss of image details and results in enhanced robustness~\cite{unser_mr}. \\

\begin{figure}[h]
\caption{Coarsest-to-Finest Image Representation-Gaussian Pyramid}
\begin{center}
\includegraphics[scale=0.45]{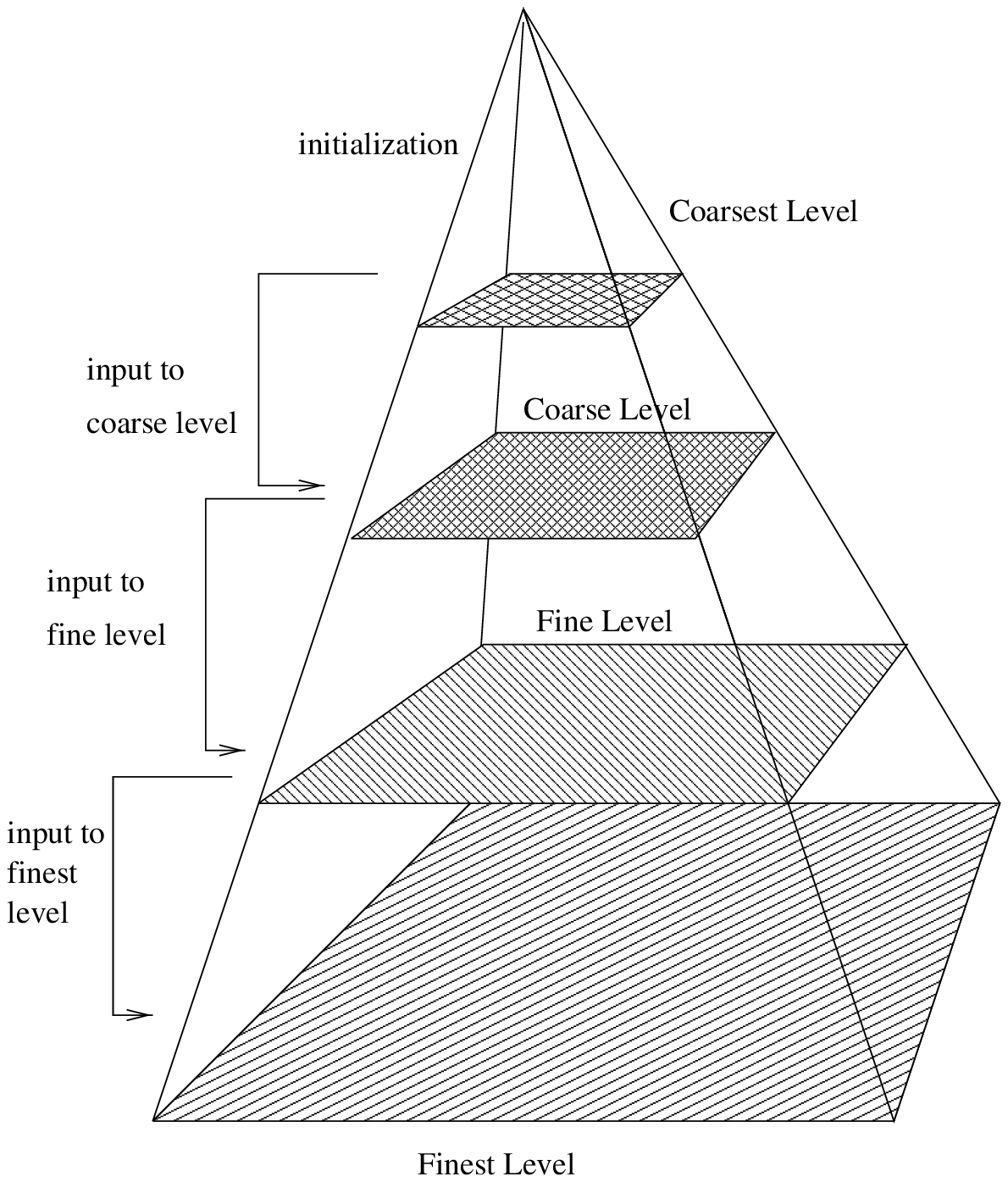} 
\end{center}
\end{figure} 

As seen from Figure 5, registration in finer level is performed with the result of the previous level as initial condition. This process continues until the finest level is reached. Since the number of iterations performed at the finest level is more relevant than the other levels, for the computational cost of the whole optimization, it is very important that the initial condition for this last level be the best possible in order to reduce the amount of refinement necessary to reach convergence. To get optimal starting conditions, it is crucial that the coarse levels of the pyramid most represent the finest level~\cite{unser_mr}. In addition to this, the use of interpolation models (except linear interpolation) will change the intensity histogram of image after each warping. Available interpolation methods vary in their computational complexity, speed and accuracy. To ensure a more accurate solution, we perform standardization after each warping/interpolation. Either small or large, intensity changes caused by the interpolation are captured by standardization.

\section{Experiments and Results}
\label{sec:results}
The registration algorithm is tested in two different experiments. The first experiment is global affine registration and the second is elastic registration, both in the multiresolution framework. Accuracy of registration results is evaluated quantitatively and qualitatively. Quantitative evaluation includes mean square differences in intensity between the pair of images \footnote{Mean Square Error}. For visual assesment of registration, a checkerboard image is created by picking alternating squares from the image pair, the registered source and target. Spatial alignment can be investigated visually by looking at continuity of tissue boundaries in the two interleaved images. The alignment is good if there is no discontinuity of contours and the tissues merge smoothly across the checkerboard borders. 

\subsection{Experiment I}
In the first experiment global deformations are captured with affine transformations in the multiresolution framework. For both the source and target images, four level coarsest to finest Gaussian pyramid is used. In  each multiresolution level, an optimal solution is determined and used as the starting point for the next level as seen in Figure 5. In each level, a single global affine transform is estimated with domain $D$, which is defined to be entire image. Transformed images are interpolated with cubic splines. Since intensities of images are changed in each warping/interpolation pair, image intensities are re-standardized into standard scale from image scale.

The algorithm is performed both on image scale and on standard intensity scale. Thirty deformed brain images are generated by applying random deformations to the original target images. Original source images are tried to be registered to each deformed target images on standard intensity scale and image scale. Random deformations are obtained by randomly choosing parameters for affine matrix $A$. The resulting deformation field is normalized so that r.m.s displacement is at most 12 pixels. An example for affinely warped images is shown in Figure 6.

\begin{figure}[h]
\begin{center}
\caption{Images are randomly deformed by affine transformations}
\includegraphics[scale=0.37]{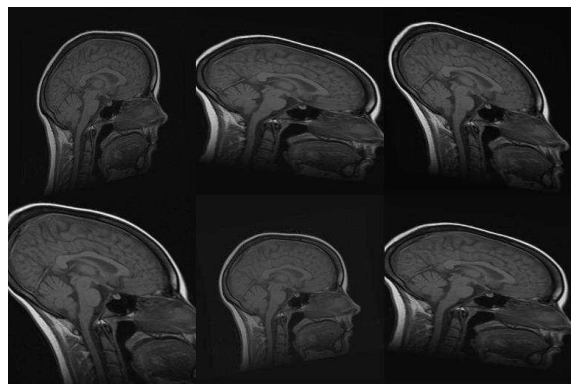} 
\end{center}
\end{figure} 

In Figure 7, three example registration results of randomly and affinely warped images are shown. The resulting images clearly show that registered source images are in good agreement with target images. Registration quality is measured over 30 randomly deformed images by mean of the square of the differences in intensity (MSE). Experiment has been done both in image scale and on standard scale to show improvement in MSE sense. Table-1 shows the MSE, maximum MSE and minimum MSE over 30 registration examples on image scale and on intensity scale respectively.

\begin{figure}[h]
\begin{center}
\caption{\noindent Resulting registration of images with random affine warps. Each row includes source, target and registered source}
\includegraphics[scale=0.40]{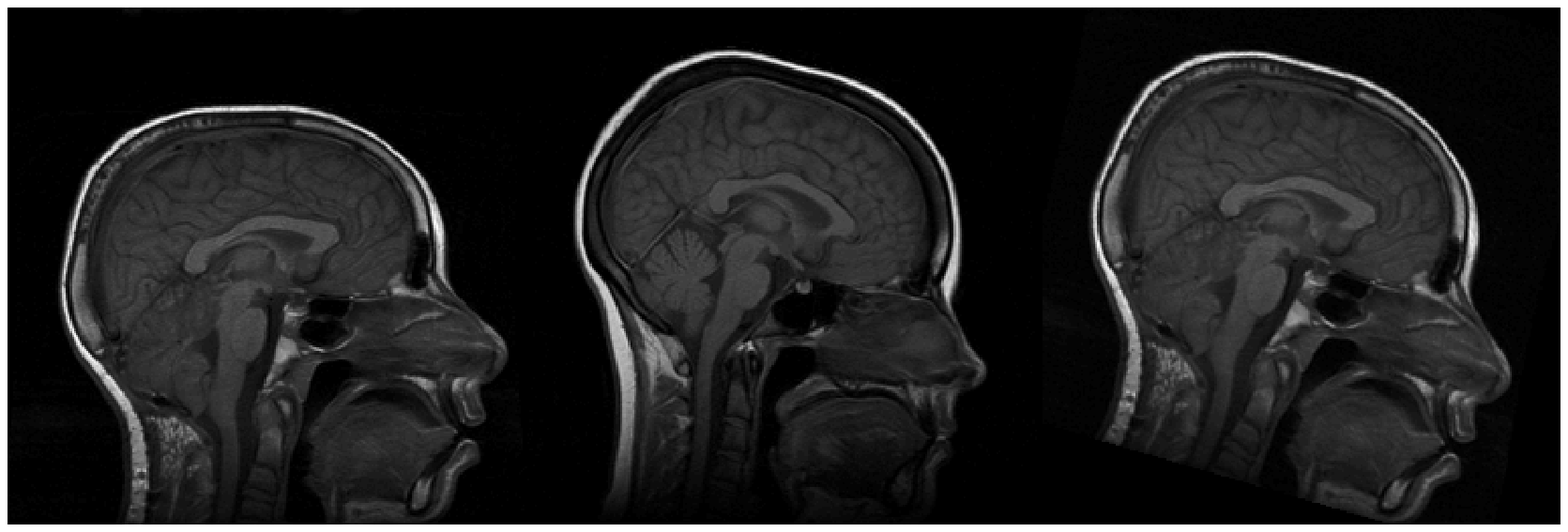}
\includegraphics[scale=0.40]{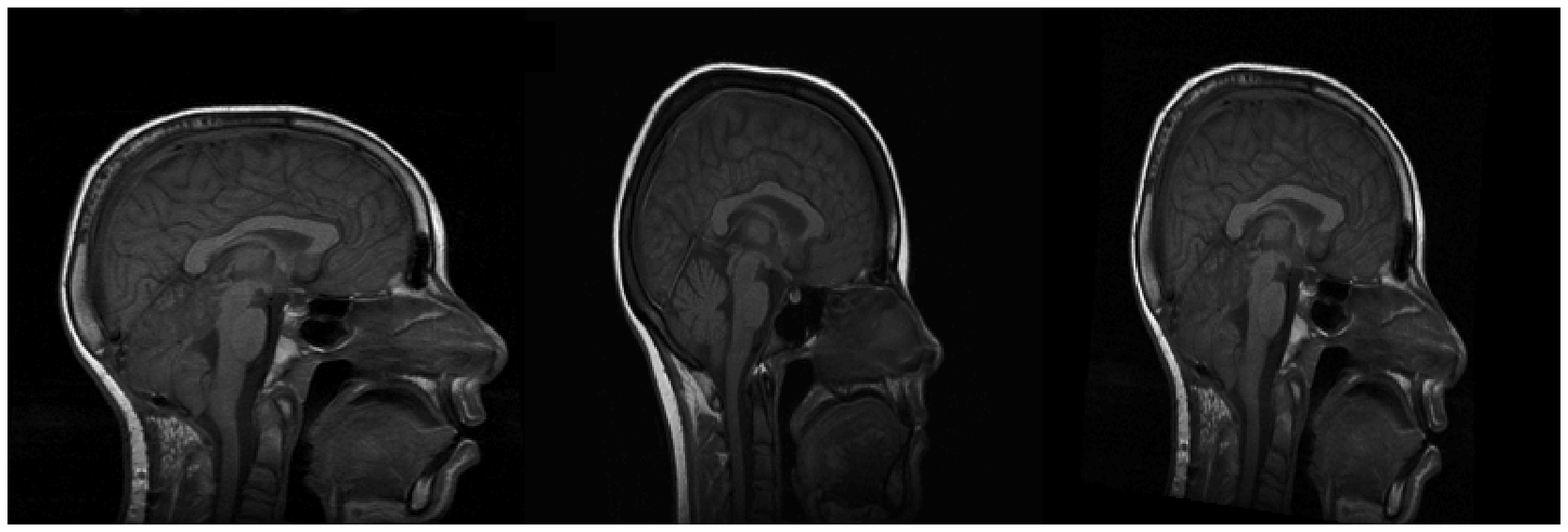}
\includegraphics[scale=0.40]{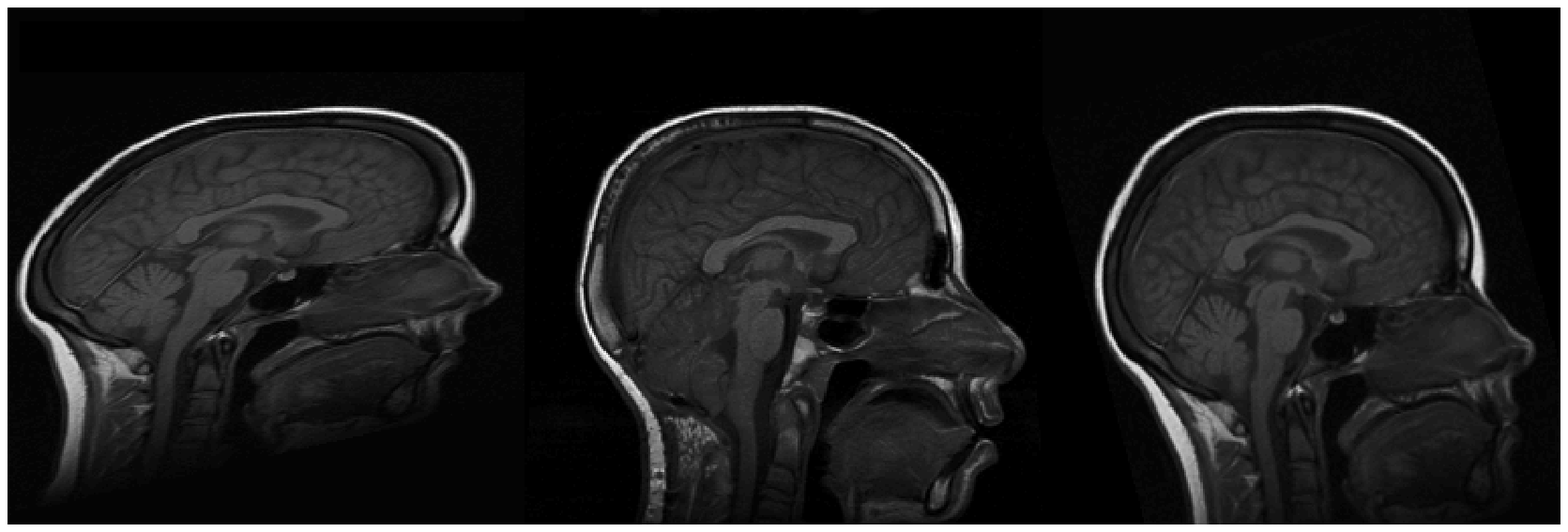}
\end{center}
\end{figure}

\begin{table}[h]
\begin{center}
\caption{Affine registration evaluations \label{resultglobal}}
\begin{tabular}{|c|c|c|c|}
\hline
Scale/Error & MSE & Max MSE & Min MSE \\ \hline
On Image Scale                    & 0.1211     & 0.1703 & 0.0993 \\  \hline
On Standard Scale                &  0.1060    & 0.1457 & 0.0857  \\ \hline
\end{tabular}
\end{center}
\end{table}

\subsection{Experiment II}
In the second experiment, in addition to global deformations, local deformations are captured with locally affine transformations in the multiresolution framework. The experiment starts with global affine registration in which pre-alignment of the source and target images is obtained. Similar to the first experiment, standardization is performed to handle image intensity variations in every warping/interpolation pair. The resulting global affine parameters are used in the coarsest level of the Gaussian image pyramid to estimate local affine parameters on domain D, where D is now 5x5 pixels. Estimated local affine parameters are used to transform source image in the next level of the Gaussian image pyramid. Accumulating all the transformations on each Gaussian image pyramid level yields a single final transformation that is able to capture both large and small scale motions.

As in the previous experiment, the algorithm is performed both on image scale and on standard intensity scale. Thirty deformed brain images are generated by applying random nonlinear deformations to the target images. Random nonlinear deformations are obtained according to equation given below:

\begin{eqnarray}
x'=n_1.x+(-1)^{n_2}.e^{n_3}.sin(y/n_4) \nonumber  \\
y'=n_5.x+(-1)^{n_6}.e^{n_7}.cos(y/n_8)
\end{eqnarray}
where $[x, y]$, $[x', y']$ are initial and transformed spatial coordinates of the images respectively, and, $n_1,..n_8$ are chosen randomly such that r.m.s displacement is at most 12 pixels. An example for nonlinearly warped images is shown in Figure 8.

\begin{figure}[h]
\begin{center}
\caption{Images are randomly deformed by nonlinear transformations}
\includegraphics[scale=0.43]{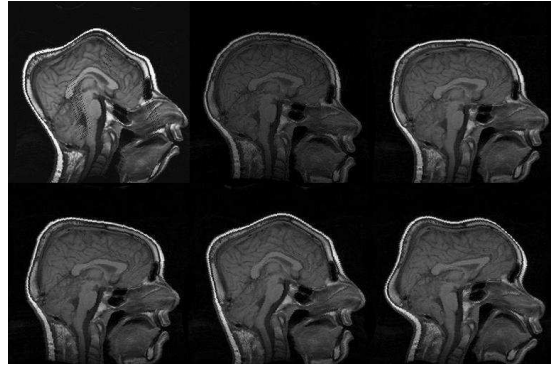} 
\end{center}
\end{figure}

In Figure 9, three example registration results of randomly and nonlinearly warped images are shown. Capturing signal intensity variations during registration process with intensity standardization method leads to assesment of visual comparision of registered source and target images with warping grid. Evaluation of the registration results is summarized in Table-2. The table shows that large and small scale deformations are captured accurately on the standard intensity scale. Resulting images have fixed intensity meanings even there is large intensity variations initially. 

\begin{figure}[h]
\begin{center}
\caption{\noindent Resulting registration of images with random nonlinear warps. Each row includes source, target, registered source and estimated warping grid}
\includegraphics[scale=0.37]{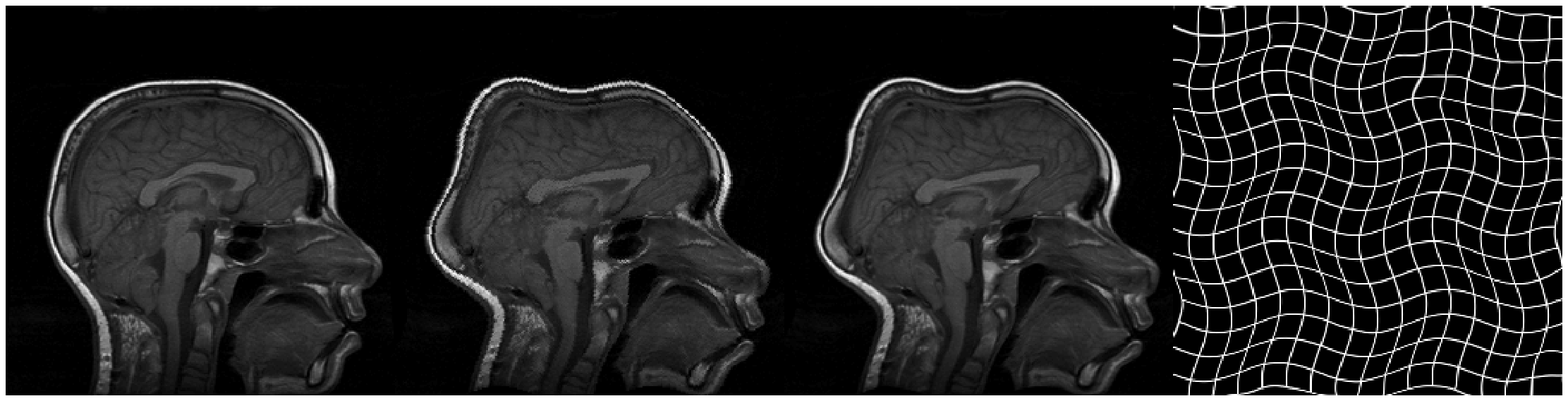}
\includegraphics[scale=0.37]{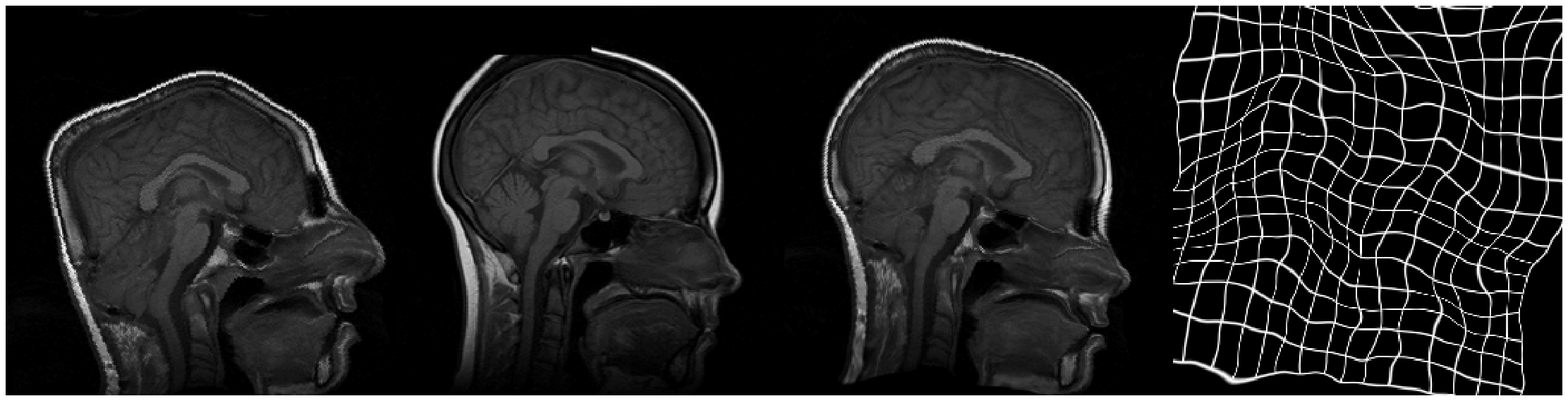}
\includegraphics[scale=0.37]{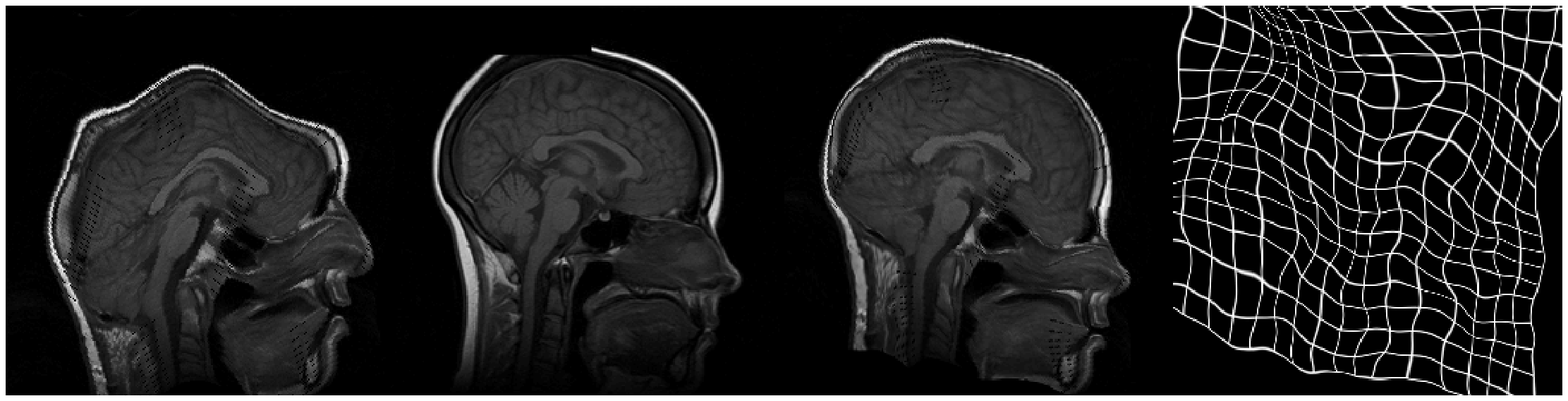}
\end{center}
\end{figure}

The resulting registered images and deformation fields show that standardization of intensity scales improves the accuracy of registration. 

\begin{table}[h]
\begin{center}
\caption{Elastic registration evaluations \label{resultlocal}}
\begin{tabular}{|c|c|c|c|}
\hline
Scale/Error & MSE & Max MSE & Min MSE \\ \hline
On Image Scale                    &  0.0204     & 0.0356 & 0.0155 \\  \hline
On Standard Scale                &  0.0194    & 0.0320 & 0.0148  \\ \hline
\end{tabular}
\end{center}
\end{table}

Another method to evaluate proposed registration method is visual examination of checkerboard images. Figure 10 shows an elastic registration example together with checkerboard image illustrating how well the image pair is registered. Checkerboard image includes white and black squares corresponding to intensity values taken from the registered source and the target image respectively. Our overall observation from experimental results is that multiresolution elastic registration on standard intensity scale can capture both local and global deformations with high accuracy. 

\begin{figure}[h]
\begin{center}
\caption{\noindent  First row includes the source and target images, second row shows registered source with warped grid and checkerboard image for visual assesment}
\includegraphics[scale=0.70]{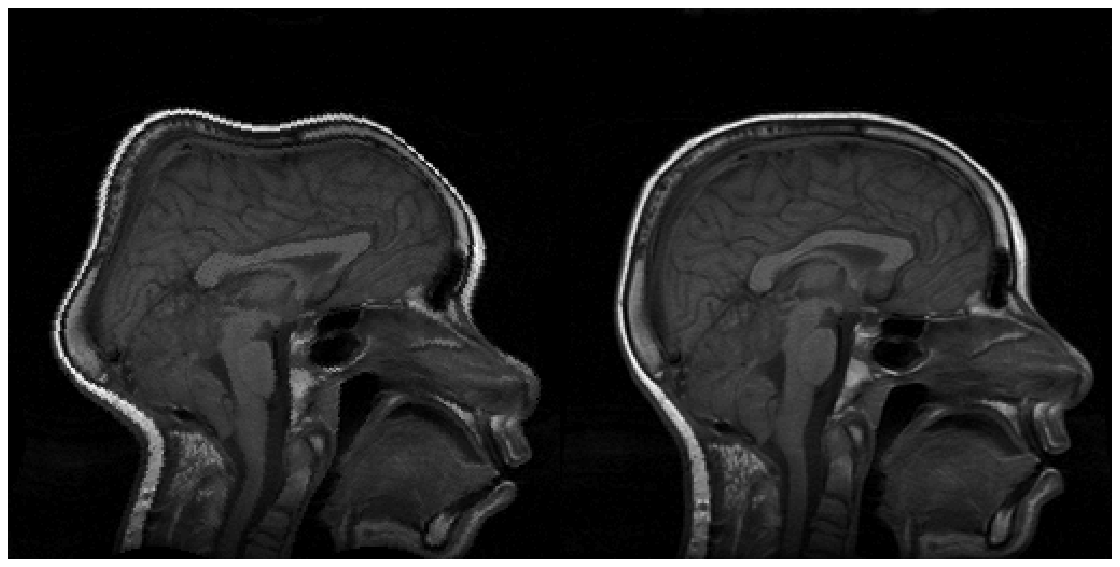}
\includegraphics[scale=0.70]{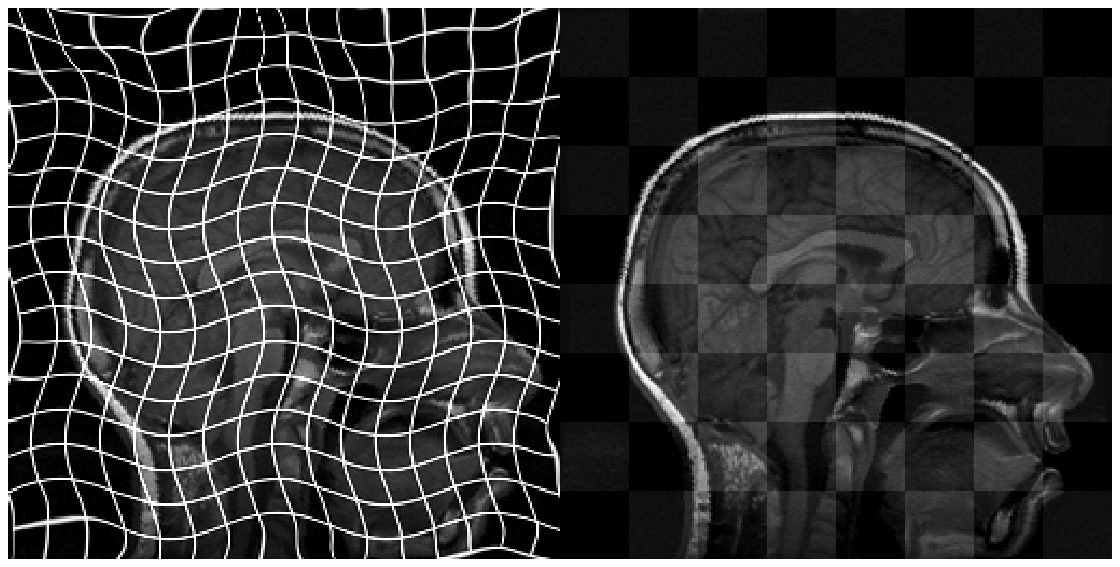}
\end{center}
\end{figure}

\section{Conclusion}
\label{sec:conc}
In this paper, a multiresolution elastic medical image registration is presented. The multiresolution framework leads to a robust and fast algorithm. Variations in image intensity histograms at each pyramid level is corrected by the intensity standardization method in order to provide an efficient framework for registration. The standardization method enables similar image intensities mean similar tissue contents, which leads to less parameters in registration. Qualitative and quantitative evaluation of experimental results indicate that small and large scale deformations are captured with high accuracy. The examples presented here use 2D images but the proposed algorithm is valid in 3D as well.

\section{Acknowledgements}
This research is funded by the European Commission Fp6 Marie Curie Action Programme (MEST-CT-2005-021170) 




\bibliographystyle{latex8}
\bibliography{31876}

\end{document}